\documentclass{iet-ell}
\usepackage{graphicx}
\usepackage{amsmath} 
\usepackage{color}
\usepackage{todonotes}

\usepackage{algorithm}
\usepackage{algpseudocode}  
\usepackage{multirow}
\usepackage{booktabs}
\usepackage{subfigure}
\usepackage{hyperref}

\setcopyright{open}
\ietVolume{00}
\ietYear{2021}
\ietDoi{ell.10001}

\received{DD MMMM YYYY}
\accepted{DD MMMM YYYY}

\begin{document}

\title{Learning Person-specific Network Representation for Apparent Personality Traits Recognition}

\author[af1]{Fang Li}
\orcid{1234-5678-9012-3456}

\affil[af1]{School of communication and information engineering,\\
 Shanghai Technical Institute of Electronics and Information\\}

\corresp{Email: lf1266@163.com}

\begin{abstract}

\noindent Recent studies show that apparent personality traits can be reflected from human facial behavior dynamics. However, most existing methods can only encode single-scale short-term facial behaviors in the latent features for personality recognition. In this paper, we propose to recognize apparent personality recognition approach which first trains a person-specific network for each subject, modelling multi-scale long-term person-specific behavior evolution of the subject. Consequently, we hypothesize that the weights of the network contain the person-specific facial behavior-related cues of the subject. Then, we propose to encode the weights of the person-specific network to a graph representation, as the personality representation for the subject, allowing them to be processed by standard Graph Neural Networks (GNNs) for personality traits recognition. The experimental results show that our novel network weights-based approach achieved superior performance than most traditional latent feature-based approaches, and has comparable performance to the state-of-the-art method. Importantly, the produced graph representations produce robust results when using different GNNs. This paper further validated that person-specific network's weights are correlated to the subject's personality.

\end{abstract}

\maketitle


\section{Introduction}
Video-based apparent personality recognition aims to automatically infer personality traits (e.g., the Big-Five personality traits \cite{mccrae1987validation}) from human non-verbal visual spatio-temporal behaviors (i.e., behavior videos). It has drawn more attentions in recent years since this technique can benefit various real-world applications, including health diagnosis \cite{jaiswal2019automatic}, employment \cite{eddine2017personality}, etc. 

To predict a subject's personality traits from a video, existing approaches usually feed each frame or short video segment of the video \cite{wei2018deep,celiktutan2017automatic,zhang2019persemon,liao2022open} or the down-sampled video \cite{li2020cr} to the ML model, and generate frame/segment-level predictions or video-level predictions. While personality traits represent human aspects that are stable over time but differ across individuals \cite{kassin2003essentials}, frame/short segment-based approaches failed to model such long-term personalised behaviors, while down-sampling videos would lose short-term human behavior details, both of which are crucial to reflect personality. 

Inspired by recent works \cite{song2022learning,shao2021personality,song2021self} which found that parameters of person-specific ML models are highly correlated with the corresponding subject's personality, this paper proposes a novel self-supervised learning algorithm to learn a person-specific network for each subject, aiming to jointly model both short-term behavior details and long-term behaviors that are stable over time but differ across individuals. Then, we propose to encode the unique weights of each person-specific network into a graph, allowing the network weights to be parameterized as a learnable representation that can be processed by standard Graph Neural Networks (GNN) for personality recognition.

The main novelties of our approach reside in: 1. we propose to use network's weights as the representation for personality recognition; 2. we propose a novel self-supervised learning algorithm to model multi-scale person-specific behavior evolution which differs from \cite{song2021self}, where each person-specific network only models short-term facial dynamics; and 3. we propose a novel graph encoding algorithm that can encode all weights of the learned network into a standard graph representation for down-stream tasks.

\begin{figure*}[htbp]
\centerline{\includegraphics[scale=0.7]{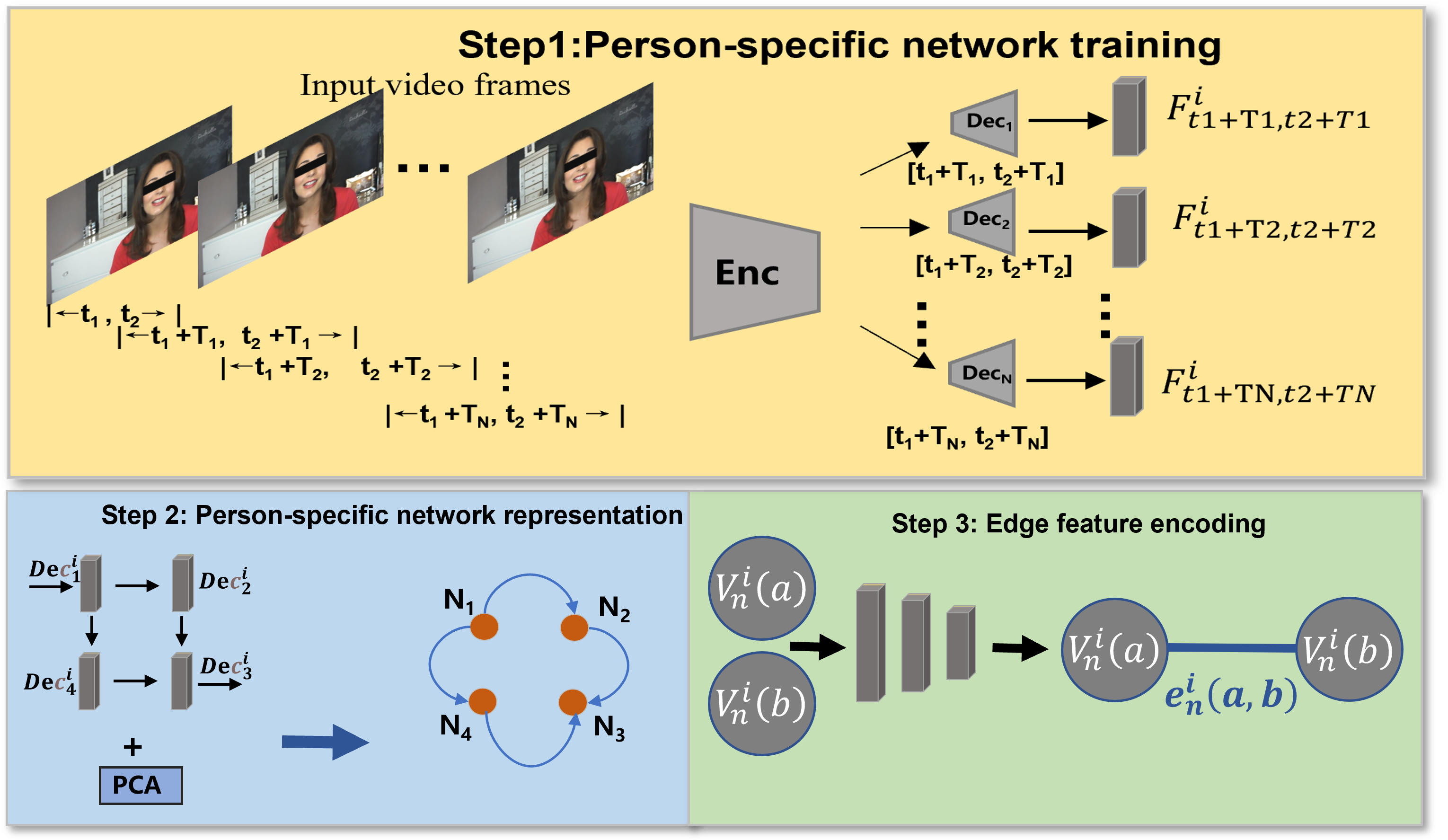}}
\caption{The pipeline of the proposed approach.}
\label{pipeline}
\end{figure*}

\section{Methodology} The proposed approach consists of two main steps: 1. person-specific network training that individually trains a network for each subject, modelling multi-scale, person-specific and personality-related behavior evolution of the subject; and 2. network graph representation generation that encodes unique and person-specific weights of each obtained person-specific network into a graph.

\subsection{Person-specific network training} As shown in Fig. \ref{pipeline}, the proposed person-specific network (P-Net) consists of an encoder \textit{Enc} and multiple decoders \textit{Dec$_1$, Dec$_2$, $\cdots$, Dec$_N$}. Specifically, the encoder takes a short video segment at the period $[t_1, t_2]$ in the video as the input, and aims to predict multiple succeeding spatio-temporal behaviours at periods $[t_1+T_1, t_2+T_1], [t_1+T_2, t_2+T_2], \cdots, [t_1+T_N, t_2+T_N]$ in the video, respectively. This process can be formulated as:
\begin{equation}
    F_{t_1+T_1, t_2+T_1}, \cdots, F_{t_1+T_N, t_2+T_N} = \text{P-Net}(F_{t_1, t_2})
\end{equation}
Specifically, we follow the similar strategy of the CR-Net \cite{li2020cr} to down-sample each video, and then train the encoder using all videos in the training set, i.e., the encoder predicts video-level apparent personality traits from each down-sampled video. In this paper, we take the latent feature maps $L_{t_1, t_2}$ produced by the last convolution layer as the output, which is denoted as
\begin{equation}
    L_{t_1, t_2} = \text{Enc}(F_{t_1, t_2})
\end{equation}
This supervised training step leads latent features extracted by the encoder to contain personality-related cues. As a result, decoders are specifically learned to process personality-related features. We hypothesize that this enforces their weights also to contain more personality-related cues (validated in Sec. \ref{sec: experiment}).


Then, for each subject $s_i$ in both training and inference stage, we individually train a set of person-specific decoders $\text{Dec}_n^i$ ($n = 1, 2, \cdots, N$) in an self-supervised manner, which are learned to model how the spatio-temporal facial behaviors $F_{t_1+T_n, t_2+T_n}^i$ at periods $[t_1+T_n, t_2+T_n]$ are evolved from the facial behavior $F_{t_1, t_2}^i$ at the period $[t_1, t_2]$. This can be formulated as:
\begin{equation}
\begin{split}
    F_{t_1+T_1, t_2+T_1}^i &= \text{Dec}_1^i(L_{t_1, t_2}^i) \\
    F_{t_1+T_2, t_2+T_2}^i &= \text{Dec}_2^i(L_{t_1, t_2}^i) \\
    &\cdots \\
    F_{t_1+T_N, t_2+T_N}^i &= \text{Dec}_N^i(L_{t_1, t_2}^i)    
\end{split}
\end{equation}
As a result, these person-specific decoders $\text{Dec}_1^i, \text{Dec}_2^i, \cdots, \text{Dec}_1^N$ are optimized to model behavior evolution of $N$ different time scales ($T_1, T_2, \cdots, T_N$) for the subject $s_i$, i.e., the weights of these decoders are adapted to the person-specific behavior evolution of the subject $s_i$.

In this paper, we choose the ResNet-17 as the encoder, where the last FC layer is removed. It takes a sequence of continuous face frames at the period $[t_1, t_2]$ as the input and generate a set of latent feature maps. Then, all decoders (three decoders are used in these paper) having the same architecture, i.e., each decoder consists of five blocks, where each block consists of a convolution layer, a ReLU activation function and an up-sampling layer. Each decoder individually takes the latent feature maps produced by the encoder, and aims to re-produce a sequence of face frames at the period $[t_1+T_n, t_2+T_n]$.

\subsection{Person-specific network representation generation} Once the decoders of person-specific network are well trained (overfitting to the training video) for the subject $s_i$, we encode its person-specific weights as a graph $\mathcal{G}^i$:
\begin{equation}
    \mathcal{G}^i(V,E) = \text{G}(\text{Dec}_1^i, \text{Dec}_2^i, \cdots, \text{Dec}_N^i)
\end{equation}
where $V \subseteq \{\mathbf{v(a)} \in \mathbb{R}^{1 \times K} \}$ is a set of vertices, and $E \subseteq \{ \mathbf{e_{i,j}} = (\mathbf{v(a)}, \mathbf{v(b)}) \mid \mathbf{v(a)}, \mathbf{v(b)} \in \mathcal{V},  a \neq b \}$ denotes a set of edges that connect vertices. In this paper, we treat each layer in decoders as a vertex in the graph. Specifically, for a layer (e.g., convolution layer, FC layer, etc) that has learnable weights, we first concatenate all weights of the layer as a vector, and then apply PCA to reduce the vector as a $C$-dimensional vertex feature. We also artificially create a unique $C$-dimensional vertex feature (e.g., $[1,1, \cdots, 1]$ and $[2,2, \cdots, 2]$) to represent each type of layers that do not have learnable weights (e.g., up-sampling, pooling, etc.). These process is represented as:
\begin{equation}
\begin{split}
    \mathbf{v}^i_n(a) = 
    \begin{cases}
    \text{PCA}(W^i_n(a)) & \text{if} \quad W^i_n(a) = \emptyset\\
    [j,j, \cdots, j] & \text{if} \quad W^i_n(a) \neq \emptyset
    \end{cases}    
\end{split}
\end{equation}
where $W^i_n(a)$ denotes the learnable weights of the $a_\text{th}$ layer in the decoder $\text{Dec}_n^i$. Consequently, all weights of person-specific decoders are contained in the graph. It should be noted that for each subject, all person-specific decoders are learned in an self-supervised manner without requiring any personality labels. Consequently, we learn a person-specific network for each subject at both training and inference stages. Meanwhile, we define if a layer's (corresponded to the vertex $\mathbf{v}^i_n(a)$) output is the input of another layer (corresponded to the vertex $\mathbf{v}^i_n(b)$), there exists an edge $\mathbf{e}^i_n(a,b)$ in the graph $\mathcal{G}^i(V,E)$ to connect $\mathbf{v}^i_n(a)$ and $\mathbf{v}^i_n(b)$. In addition, we follow the same strategy as \cite{song2022gratis} to learn a multi-dimensional feature for each existed edge under the supervision of personality labels, describing the rich personality-related relationship information between each pair of connected vertices, which can be formulated as
\begin{equation}
    \mathbf{e}^i_n(a,b) = \text{ERN}(\mathbf{v}^i_n(a), \mathbf{v}^i_n(b))
\end{equation}
As a result, the final produced $\mathcal{G}^i(V,E)$ can be represented as:
\begin{equation}
\begin{split}
    \mathcal{G}^i(V,E) = & \\
    &\{ V = \{\mathbf{v}^i_1(1), \mathbf{v}^i_1(2), \cdots, \mathbf{v}^i_N(1), \cdots \}, \\
    & E = \{\mathbf{e}^i_1(1,2), \mathbf{e}^i_1(2,3), \cdots, \mathbf{e}^i_N(1,2), \cdots \}  \}
\end{split}    
\end{equation}
where $\mathbf{v}^i_n(a) \in \mathbb{R}^{1 \times K}$ and $\mathbf{e}^i_n(a,b) \in \mathbb{R}^{1 \times K}$ have the same dimensionality.

\subsection{Personality recognition using network representations} Once the graph representation $\mathcal{G}^i(V,E)$ of the person-specific network is generated, we feed it to a Graph Neural Network (GNNs) to jointly predict five personality traits of the subject $s_i$, which can be denoted as:
\begin{equation}
    P_\text{personality}^i = \text{GNN}(\mathcal{G}^i(V,E))
\end{equation}

\section{Experiments}
\label{sec: experiment}


\subsection{Dataset} We evaluate the proposed approach based on the ChaLearn First Impression database \cite{ponce2016chalearn}. It contains $10,000$ short audio-visual clips (each clip lasts $15$ seconds) recorded from $2,764$ YouTube users, which is divided to 6000 training clips, 2000 validation clips and 2000 test clips. Each clip is annotated with Big-Five personality traits labels by multiple human annotators, with each trait ranges from 0 to 1.

\subsection{Implementation details} \textbf{Training details:} The general encoder is trained using training clips and evaluated on the validation set, based on which we then individually learn person-specific decoders for each clip for all three sets. For encoder and decoders' training, we use Adam optimizer with initial learning rate of 0.005 and 0.001, respectively. Particularly, the decoders of all subject are trained using the same strategy and initial weights, and we feed clips from the beginning of the video to the video's end, without shuffling them during the training. \textbf{GNN settings:} In this paper, two widely-used GNNs: GatedGCN \cite{bresson2017residual} and Graph Attention Network (GAT) \cite{velivckovic2017graph} are applied to process the produced network graph representations, where the employed GatedGCN consists of 5 layers and the GAT consists of 3 layers. \textbf{Metrics:} The ACC \cite{ponce2016chalearn} that has been widely used in personality computing studies is employed as the metric to evaluate the performance of listed approaches.

\subsection{Results and discussion} We first compare our approach with existing personality computing approaches in Table \ref{tb:sota}. Firstly, it can be seen that our approach has clear advantages over the existing frame/short segment-level approaches, with at least $0.6\%$ ACC improvements, demonstrating that our approach can additional encode long-term personality-related behavior information from videos. Particularly, we found our approach also generated better results than PAL \cite{song2021self} for four traits, which also learns a person-specific layer by modelling short-term facial dynamics for each subject. This may indicate that our multi-scale long-term behavior modelling scheme is a superior algorithm to learn person-specific networks. Finally, our approach has comparable performance to the state-of-the-art method \cite{li2020cr}, providing a strong evidence that the weights of the network that adapts to person-specific behaviour are also highly correlated to the subject's personality.

Table \ref{tb:ablation} displays results of a set of ablation studies. It is clear that directly training an encoder (ResNet) to make frame-level predictions and averaging them does not provide decent video-level personality predictions. It also can be observed that the proposed graph representation can much better represent the personality-related information contained in the learned decoders' weights, as all graph representation-based systems outperformed the system Vec (Dec-M). Since the GAT (Dec-M) produced better results on all five traits, it can be concluded that: 1. graph representations of multiple decoders, which contains multi-scale behavior evolution cues, contain more personality-related information, i.e., GAT (Dec-M) outperforms GAT (Dec-S); 2. Pre-training of the encoder is crucial, as GAT (Dec-M) achieved more than $0.4\%$ average ACC improvement than GAT (Dec-M-NPT); and 3. Both GAT and GatedGCN provide good performance, further suggesting that the proposed network graph representation contain crucial personality-related cues.

\setlength{\tabcolsep}{2pt}
\begin{table}[t!]
	\begin{center}
\resizebox{1\linewidth}{!} {
		\begin{tabular}{|l| c c c c c c|}
			\toprule
Methods & Extra & Agree & Consc & Neuro & Open & Avg.   \\
\hline \hline

Baseline \cite{escalante2020modeling} & 0.9019 & 0.9059 & 0.9073 & 0.8997 & 0.9045 & 0.9039  \\

PML \cite{bekhouche2017personality}  & 0.9155 & 0.9103 & 0.9137&  0.9082& 0.9100 &0.9115  \\

NJU-LAMDA  \cite{wei2018deep} & 0.9112 & 0.9135 & 0.9128 & 0.9098 & 0.9105 & 0.9116  \\

DCC \cite{guccluturk2016deep}  & 0.9088 & 0.9097 & 0.9109 & 0.9085 & 0.9092 &0.9109  \\

\hline

CR-Net \cite{li2020cr} & \textbf{0.9200} & 0.9176 & \textbf{0.9218} & [0.9150] & [0.9191] & \textbf{0.9187} \\

PerEmoN \cite{zhang2019persemon} & \textbf{0.920} & 0.914 & [0.921] & 0.914 & 0.915 & 0.917 \\

DS \cite{li2022domain} & 0.9138 & 0.9190 & 0.9166 & 0.9123 & \textbf{0.9198} & 0.9163 \\

\hline

PAL \cite{song2021self} & 0.9183 & 0.\textbf{9262} & 0.9082  & 0.9133 & 0.9180 & 0.9168  \\

Ours  & 0.9141 & [0.9218] & 0.9197 & \textbf{0.9165} & 0.9152 & [0.9174] \\

			\bottomrule
		\end{tabular}
        }
	\end{center}
	\caption{ACC results achieved by the proposed approaches and existing approaches, where the bold numbers denote the best result and bracketed numbers denote the second best result.}  
\label{tb:sota}
\end{table}
\setlength{\tabcolsep}{1.4pt}

\setlength{\tabcolsep}{2pt}
\begin{table}[t!]
	\begin{center}
\resizebox{1\linewidth}{!} {
		\begin{tabular}{|l| c c c c c c|}
			\toprule
Methods & Extra & Agree & Consc & Neuro & Open & Avg.   \\
\hline \hline

Encoder & 0.8981 & 0.9068 & 0.9021 & 0.9007 & 0.9015   & 0.9018  \\

Vec (Dec-M) & 0.8976 & 0.8991 & 0.9086 & 0.9012 & 0.9024   & 0.9018  \\

GatedGCN (Dec-M) & 0.9159 & 0.9213 & 0.9174 & 0.9128 & 0.9177 &   0.9170  \\

GAT (Dec-S)   & 0.9130 & 0.9186 & 0.9166 &0.9171 & 0.9138 &  0.9158  \\

GAT (Dec-M-NPT) & 0.9098 & 0.9185 & 0.9137 & 0.9154 & 0.9107   & 0.9136  \\

GAT (Dec-M) & 0.9141 & 0.9218 & 0.9197 & 0.9165 & 0.9152 & 0.9174 \\

			\bottomrule
		\end{tabular}
        }
	\end{center}
	\caption{ACC results achieved by different setting of the proposed approach, where \textit{Dec-S}, \textit{Dec-M}, and \textit{Dec-M-NPT} denotes the use of a single decoder, multiple decoders and multiple decoders trained without the pre-training of the encoder. \textit{Vec} denotes that we only concatenate all weights of the decoders as a vector and feed it to MLP for personality recognition.}  
\label{tb:ablation}
\end{table}
\setlength{\tabcolsep}{1.4pt}

\section{Conclusions}

\noindent This paper presents a novel automatic personality recognition approach that trains a person-specific network for each subject, which models the unique long-term facial behavior evolution of the subject in a self-supervised way. Then, we encode the weights of each person-specific network as a graph personality representation which contain the behavior features obtained from the whole video. While our approach is one of the first attempt that uses network's weights to represent human behaviors, it achieved better performance to well-developed traditional deep learning approaches, and comparable results to the state-of-the-art method, showing that the weights of the CNN trained with person-specific behaviors are correlated with the corresponding person's personality. In particular, our approach has clear advantages over an frame-level method which also uses network's weights as the person-specific representation. This indicates that the proposed long-term facial behavior evolution modelling approach is superior than the short-term facial dynamics modelling method, i.e. our method is the best scheme to encode personality cues into network's weights. The main limitation of our approach is that it requires to learn a network for each subject and thus is time-consuming. Our future work would focus on addressing this problem by considering transfer learning or other weights sharing schemes.

\bibliographystyle{iet}
\bibliography{iet-ell}

\end{document}